\documentclass[conference]{IEEEtran}
\IEEEoverridecommandlockouts
\usepackage{cite}
\usepackage{balance}
\usepackage{amsmath,amssymb,amsfonts}
\usepackage{algorithmic}
\usepackage{graphicx}
\usepackage{flushend}
\usepackage{textcomp}
\usepackage{xcolor}
\usepackage{microtype}
\usepackage[hidelinks]{hyperref}
\def\BibTeX{{\rm B\kern-.05em{\sc i\kern-.025em b}\kern-.08em
    T\kern-.1667em\lower.7ex\hbox{E}\kern-.125emX}}

\usepackage{amsthm}

\newtheorem{definition}{Definition}

\usepackage{multirow}
\usepackage{colortbl}
\usepackage{booktabs}

\usepackage{subcaption}

\usepackage{xspace}
\newcommand{\tool}{{\small \textsf{TOWER}}\@\xspace}
\newcommand{\tower}{\tool}
\newcommand{\lime}{{\small \textsf{LIME}}\@\xspace}

\usepackage{tcolorbox}
\usepackage{xspace}
\usepackage{fontawesome}
\newcommand{\editnote}[3]{\xspace\colorbox{#1}{\textcolor{white}{~\faCommenting{}~#2~}}\textcolor{#1}{~#3}\xspace}

\newcommand{\stefano}[1]{\editnote{olive}{Stefano}{#1}}

\begin{document}

\title{Automated Trustworthiness Testing \\ for Machine Learning Classifiers}

\author{\IEEEauthorblockN{Steven Cho}
\IEEEauthorblockA{\textit{University of Auckland}\\
Auckland, New Zealand \\
scho518@aucklanduni.ac.nz}
\and
\IEEEauthorblockN{Seaton Cousins-Baxter}
\IEEEauthorblockA{\textit{University of Auckland}\\
Auckland, New Zealand \\
scou766@aucklanduni.ac.nz}
\and
\IEEEauthorblockN{Stefano Ruberto}
\IEEEauthorblockA{\textit{JRC European Commission} \\
Ispra, Italy \\
stefano.ruberto@ec.europa.eu}
\and
\IEEEauthorblockN{ Valerio Terragni}
\IEEEauthorblockA{\textit{University of Auckland}\\
Auckland, New Zealand \\
v.terragni@auckland.ac.nz}
}

\maketitle

\begin{abstract}

Machine Learning (ML) has become an integral part of our society, commonly used in critical domains such as finance, healthcare, and transportation. Therefore, it is crucial to evaluate not only whether ML models make correct predictions but also whether they do so for the correct reasons, ensuring our trust that will perform well on unseen data. This concept is known as trustworthiness in ML. Recently, explainable techniques (e.g., LIME, SHAP) have been developed to interpret the decision-making processes of ML models, providing explanations for their predictions (e.g., words in the input that influenced the prediction the most). Assessing the plausibility of these explanations can enhance our confidence in the models' trustworthiness. However, current approaches typically rely on human judgment to determine the plausibility of these explanations.

This paper proposes TOWER, the first technique to automatically create trustworthiness oracles that determine whether text classifier predictions are trustworthy. It leverages word embeddings to automatically evaluate the trustworthiness of a model-agnostic text classifiers based on the outputs of explanatory techniques. Our hypothesis is that a prediction is trustworthy if the words in its explanation are semantically related to the predicted class. 

We perform unsupervised learning with untrustworthy models obtained from noisy data to find the optimal configuration of TOWER. We then evaluated TOWER on a human-labeled trustworthiness dataset that we created. The results show that TOWER can detect a decrease in trustworthiness as noise increases, but is not effective when evaluated against the human-labeled dataset. Our initial experiments suggest that our hypothesis is valid and promising, but further research is needed to better understand the relationship between explanations and trustworthiness issues.

\end{abstract}


\section{Introduction}

Machine Learning (ML) has become integral to many areas of society. With its prevalence, it is crucial to determine whether we can \textit{trust} ML models. \emph{Trust goes beyond correctness}. An ML model can achieve high accuracy and make correct predictions yet still be untrustworthy. This occurs when the reasons behind its correct predictions are flawed, making the model unreliable for unseen data.

\smallskip
One way to measure trustworthiness is through understanding the decision-making process of ML models~\cite{Trustworthy-Explainability-Acceptance-Medical, there-is-hope-after-all, ritual}. While there are ML classifiers that are inherently explainable (e.g., decision trees), most ML classifiers (e.g., (deep) neural networks) cannot directly explain their decisions in a way that a human would understand. However, the area of \textbf{Explainable Machine Learning (XML)}~\cite{explainable-ai-meta-survey} studies  techniques to explain the predictions of any classifier, as long as it has interpretable inputs (i.e., text, numbers, or images). 

\smallskip
One of the first and most representative XML technique is \lime~\cite{lime}. It produces textual or visual artifacts for a qualitative understanding of the relationship between the test input instance’s components (e.g., words in a text) and the model’s prediction. For example, for text classification, \lime produces a list of words that most influenced the prediction. Given such explanations, a human can judge if they are coherent with the problem domain, and thus the ML model is trustworthy and will likely generalize well to unseen data.

\begin{figure}[t!]
    \centerline{\includegraphics[width=0.5\textwidth]{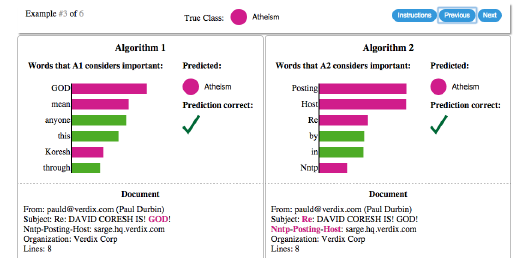}}
    \vspace{-2mm}
    \caption{Example of ``untrustworthy'' prediction from \lime paper~\cite{lime}. Algorithm 2 based the prediction on irrelevant words: “Posting”, “Host”, “Re”, and “Nntp”}
     \vspace{-2mm}
    \label{fig:lime}
\end{figure}

\smallskip
An example discussed in the \lime paper is an ML model that classifies emails belonging to “Christianity” or “Atheism” newsgroups (see Fig.~\ref{fig:lime}). Although the classifier achieves 94\% accuracy, \lime's explanations show that some predictions were made for arbitrary reasons: the words “Posting”, “Host”, and “Re” were prevalent in only one class and have nothing to do with either “Christianity” or “Atheism” (see algorithm 2 in Fig.~\ref{fig:lime}). While it is easy for a human to conclude that these predictions are not trustworthy, relying on the manual analysis of classifier explanations is very costly.

% The most common forms of machine learning model evaluation measure the capabilities and performance of the model. This, however, is often insufficient in determining the 

%The evaluation of machine learning models is a well-researched subject, using metrics including accuracy, model relevance, efficiency, robustness, fairness, and interpretability~\cite{machine-learning-testing-survey-landscapes-and-horizons}. However, much less explored is the space of determining, quantifiably, how far a model can be trusted -- that is, measuring a model's \textit{trustworthiness}.

% add a motivating example here?

%One method is through understanding the decision-making process of models. The field of explainable machine learning (XAI) explores this through the use of various different techniques~\cite{explainable-ai-meta-survey}. There have been a number of proposals using these techniques and others to measure the trustworthiness of a model~\cite{Trustworthy-Explainability-Acceptance-Medical, there-is-hope-after-all, ritual}. These methods, however, are slow and potentially expensive -- they all require human involvement in order. There has been seemingly no attempt at creating an automated method to quantify the trustworthiness of a model.

\smallskip
This paper presents \textbf{T}rustworthiness \textbf{O}racle through \textbf{W}ord \textbf{E}mbedding \textbf{R}elatedness (\textbf{\tower}), \textbf{the first automated trustworthiness testing technique for ML text classifiers}. 
\tower generates automated oracles that replace a human in judging if the explanations of a ML prediction are trustworthy or not. The key advantage of an automated approach is to avoid the manual cost of inspecting XML's explanations. Manual judgement (as envisioned by Ribeiro et al. when proposing \lime) is infeasible to perform for many ML predictions.

\smallskip
\tower is based on the intuition that such trustworthy oracles can be automatically constructed by relying on \textbf{word embedding models}. Embedding models use vectors to represent words in a low-dimensional space to quantify semantic relatedness between words. All these models have in common that two words are semantically related if the vectors representing them are close in the vector space. The inverse of such a distance is the semantic similarity score (usually [0,1]). Using word embeddings, \tower automatically checks if the explanations produced by XML techniques contain words that are semantically related to the class name/description. For example, consider the “Christianity” and “Atheism” classifier discussed above, the \texttt{word2vec-google-news-300} embedding model returns low semantic similarity scores between the \lime explanation “Post” and class labels “Christianity” and “Atheism” (0.054 and 0.065, respectively). In the context of \tower, such a low similarity score could indicate trustworthiness issues in the data or model (like in the example discussed in Fig.~\ref{fig:lime}). 

We trained TOWER's configuration parameters using unsupervised learning with around 1,600 untrustworthy and trustworthy models obtained from noisy data as a proxy for trustworthiness issues. Then, we took the learned optimal configuration of TOWER and evaluated it by creating a human-labeled trustworthiness dataset of 328 instances.

On one hand, experiments with noisy data show promising results, demonstrating that TOWER can effectively detect a decrease in trustworthiness as noise increases. On the other hand, when evaluated against the human-labeled dataset, the TOWER trained configuration exhibited low performance. Our initial experiments suggest that our hypothesis is valid and promising, but further research is needed to understand the poor performance on the human-labeled dataset. This might be because the types of noise considered are not a good proxy for real-world trustworthiness issues. Nonetheless, this paper marks the first significant step towards automated trustworthiness testing, which is an exciting new research direction.

In summary, this paper makes the following contributions:

\begin{itemize}
\item We define the concept of automated trustworthiness testing and oracles for ML text classifiers;
    \item We present \tower, the first technique to automatically determine the trustworthiness of a text classifier;
    \item We discuss a series of experiments that highlights the limitations of \tower;
    %\item We release a dataset of explanations from trustworthy and untrustworthy models obtained with noise injection.
    \item We release a ground-truth dataset of human-labelled trustworthy and untrustworthy explanations\footnote{\url{https://zenodo.org/records/11499368}}.
\end{itemize}

\section{Preliminaries and Problem Formulation}

This section formalises the trustworthiness testing problem and gives the preliminaries of this work.

\smallskip
\textbf{Trustworthiness} as a concept is relevant across many disciplines. In ML, trustworthiness is generally defined as a combination of related concepts, such as robustness, security, transparency, fairness, and safety~\cite{trustworthy-ai, trustworthy-ai-from-principles-to-practices}. Instead, in this paper, we rely on the following definition by Kästner et al.~\cite{On-the-Relation-of-Trust-and-Explainability}:

\begin{definition}
A system is \textbf{trustworthy} to a stakeholder S in a context C if the system works properly in C, and S is justified in their belief of that, given they believe it.
\end{definition}

Colloquially, \emph{a system is trustworthy if it is 'right for the right reasons'}. For ML classifiers, `right` means that the classification is correct and these 'right reasons' can be determined through \textit{prediction explanations}.

\medskip
\textbf{Explainability.} Most ML models are black-box; that is, their internal reasoning is opaque to an observer. To understand and trust these models, the fields of explainable AI (XAI) and explainable ML (XML) have emerged~\cite{explainable-ai-meta-survey}.

\smallskip
Various XML methods enables the interpretability of black-box ML models, including saliency maps, feature attribution methods, counterfactual and contrastive explanations, white-box models, and global surrogates~\cite{typology-of-the-concept-of-explanation}. We will focus on techniques that provide textual output -- specifically, feature attribution methods. These techniques identify the most important features in an input for a given prediction.

\smallskip
\lime~\cite{lime}, one of the most popular of these methods, works by creating a simplified model that approximates the behavior of a more complex model for a particular input. This is a system-agnostic post-hoc explainability technique. The output of \lime is a list of words from the input that it deems important to the prediction, along with their corresponding importance scores—weights measuring the importance of the phrase. Henceforth, we will refer to this phrase-score list as an \textit{explanation}.

\smallskip
Explanations have two properties: \textit{faithfulness}, which measures how well the explanation represents the workings of the model; and \textit{plausibility}, which measures how convincing the explanation is to a human~\cite{attention-is-not-not-explanation}. 
%It is a common sentiment that model explanations can promote trustworthiness~\cite{On-the-Relation-of-Trust-and-Explainability, mythos-of-model-interpretability}. Thus, 
Following previous work on trustworthiness~\cite{On-the-Relation-of-Trust-and-Explainability, mythos-of-model-interpretability,lime}, we use the plausibility of explanations as a proxy for measuring the trustworthiness of a prediction.

\begin{definition}
Given a ML prediction and its explanation, a \textbf{trustworthiness oracle} is a function that returns \textsf{true} if the explanation is plausible and thus the prediction is trustworthy, and \textsf{false} otherwise.
\end{definition}

\smallskip
The goal of \tower is to produce such oracles automatically by leveraging word embedding techniques to measure \textit{plausibility}.

\section{\tower}

\begin{figure*}[th!]
    \centering
    \includegraphics[width=1\linewidth]{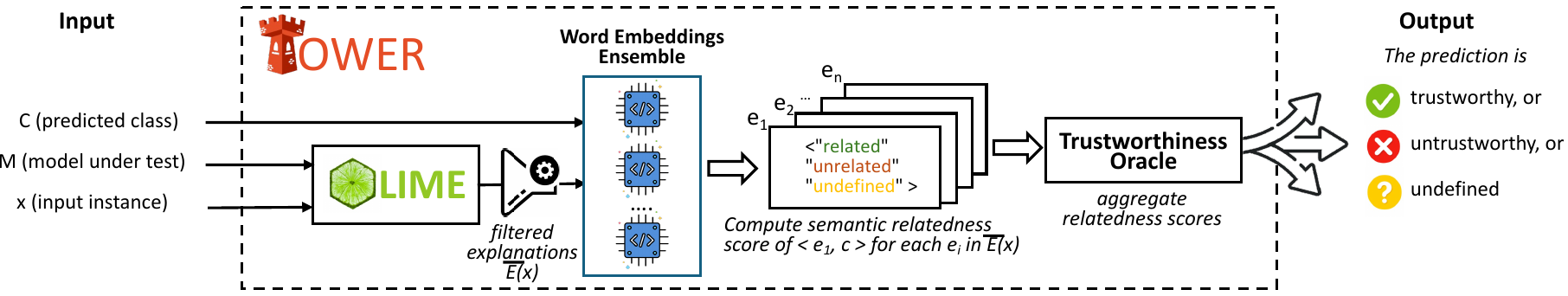}
    \vspace{-3mm}
    \caption{Logical architecture of \tower. Given an input instance $x$, a ML classifier model under test $M$, and the predicted class $c$, \tower automatically judges if the prediction $M(x) = c$ is trustworthy, untrustworthy, or undefined (non enough information/confidence to make a decision). }
    \label{fig:tower-main}
\end{figure*}

This paper presents \tower, a technique for automatically testing the trustworthiness of a given text classifier using explainability techniques and word embeddings.

Fig.~\ref{fig:tower-main} provides an overview of \tower's logical architecture. The inputs to \tower include an input instance $x$, a machine learning classifier model $M$ under test, and the predicted class $c$. \tower implements an automated trustworthiness oracle to determine if the prediction $M(x) = c$ is trustworthy, untrustworthy, or undefined (insufficient information/confidence to make a decision). Given a large number of labelled instances, we can perform trustworthiness testing and calculate the percentage of trustworthy and untrustworthy predictions. %\steven{technically it also needs the true class to determine whether the prediction is correct or not}

\smallskip
Internally, \tower uses \lime~\cite{lime} to obtain explanations for the model $M$'s prediction on $x$. It then filters out explanations with low importance scores (as measured by \lime). For each remaining explanation word $e_i \in \overline{E}(x)$, \tower relies on an ensemble of word embeddings to compute the semantic relatedness score with the predicted (and correct) class $c$. \tower aggregates the relatedness scores of all explanation words to make an informed decision about whether $M$'s prediction on $x$ is trustworthy. \tower can also respond with "undefined" if the word embeddings lack sufficient confidence about the semantic relatedness. We now discuss the internal components of \tower in more detail.
\begin{comment}

It takes as an input the model to be tested and a series of labelled test data, and outputs a trustworthiness score for the model.

\tower begins by taking as an input a labelled dataset and a model. An explanation is generated for each correctly-predicted instance of data. Each explanation is filtered to obtain the words the model considers most important to its prediction. These words are then run through a word embedding ensemble to find how related each is to the predicted class. These relatedness scores are then combined to determine a trustworthiness score for the individual prediction, and finally across all predictions for the model as a whole. Fig.~\ref{fig:tower-process} shows the overall process, and Fig.~\ref{fig:tower-main} an example.

\begin{figure}[t!]
    \centerline{\includegraphics[width=0.5\textwidth]{figures/placeholder.png}}
    \caption{The main procedure for \tower.}
    \label{fig:tower-process}
\end{figure}

\begin{figure}[t!]
    \centerline{\includegraphics[width=0.5\textwidth]{figures/diagram-main.pdf}}
    \caption{An example for the main procedure for \tower.\stefano{the last square with the average score is not clear. Why do we need an average score?}}
    \label{fig:tower-main}
\end{figure}
\end{comment}

\subsection{Input}

\tower takes as input an ML model under test, $M$, an input instance, $x$, and the predicted class, $c$. Due to the nature of \tower, it currently supports only text classification models, where the classes to be predicted are well-defined entities (e.g., “Christianity” and “Atheism”). This is because \tower needs to measure the semantic relatedness between the explanations (which must be text-based) and the predicted class (which should be a well-defined entity). More specifically, $x$ is a text, and $c$ is the predicted class to which $x$ belongs. The current version of \tower supports classes composed of a single word. The text classifier $M$ can be any black-box machine learning model that is supported by explanation techniques. This includes models obtained with popular machine learning techniques like neural networks.

\smallskip
Note that we do not consider any input instances that are not correctly predicted by the model under test, as these prediction would, by Definition 1, be untrustworthy. Thus, we assume that the class $c$ predicted by the model is the correct class for $x$. This assumption can be easily met when testing the model with labeled data, as we can automatically ignore instances that are not correctly classified.

\subsection{Explanations}

The first step of \tower uses a post-hoc black-box explanation technique to obtain the explanation of the prediction. The current implementation of \tower uses \lime~\cite{lime}, but it could be replaced with any similar technique. \tower assumes that the obtained explanation is faithful, i.e., it correctly represents the workings of the model (see Section II). This is a reasonable assumption given the good performance of \lime and similar techniques~\cite{lime}.

\smallskip
Given the model $M$ and the input instance $x$, \lime perturbs $x$ to identify the words that contributed the most to its decision-making, which constitutes the explanation. More formally, the explanation $E$ of $x$ is a list of words in $x$ with associated importance scores $s$, ordered by $s$. Formally, $E(x) = \langle (e_1,s_1), (e_2,s_2), \dots, (e_n,s_n) \rangle$, where $e_i \in x \quad \forall i = 1, \dots, n$ and $s_i > s_{i + 1} \quad \forall i = 1, \dots, n - 1$.

\smallskip
\tower has two configuration parameters to filter the words in the explanations because some of them might have low importance scores. The \textit{explanation threshold} filters out explanations with importance scores below a certain threshold, and the \textit{explanation top-n} parameter considers only the top $n$ words on the list. \tower also ignores words with negative importance scores, as they represent importance towards different classes. These are not useful for measuring the relatedness of impactful words towards the predicted class. We use $\overline{E}(x)$ to denote the filtered explanation.

\subsection{Word Embeddings Ensemble}

The second step of \tower uses word embeddings to measure the relatedness among each word in the filtered explanation $\overline{E}(x)$ and the class $c$. Word embeddings are vector representations of words, showing their semantic relationship, as found via their use in similar contexts. Two words are semantically related if their vectors are close in the vector space, typically measured using cosine similarity. The inverse of this distance is the semantic relatedness score. Our intuition is that untrustworthy predictions should contain unrelated words in their explanations.

\smallskip
Several challenges must be addressed when using word embeddings to detect untrustworthy predictions:

\textbf{1) How do we deal with word embedding models bias?} Word embedding models are trained on specific datasets, introducing biases. To mitigate this, \tower uses an \textbf{ensemble approach}, combining the results of multiple word embedding techniques trained on different datasets. This strategy should also enhance the effectiveness of the semantic similarity score computation. 

\textbf{2) How do we deal with word embedding uncertainty?} Word embeddings provide a relatedness score (usually between 0 and 1), necessitating a threshold $\tau$ to decide if the score is high enough to indicate related words. Setting a single threshold $\tau$ to differentiate trustworthy from untrustworthy explanations could result in many false positives. The semantic distances returned by word embedding models are not precise enough to ensure that a score of $\tau$ + 0.01 is trustworthy while $\tau$ - 0.01 is not. To address this, scores within a certain range of the threshold are considered 'undefined'; scores above this range are 'related' and those below are 'unrelated'. This mitigates false positives by avoiding definite decisions when the models are uncertain. The range is a configurable parameter.

\textbf{3) How do we set the thresholds?} 
Finding a single threshold that works for all embedding methods in the ensemble is challenging, as semantic relatedness scores are specific to each method and cannot be directly compared. Therefore, we propose an automated way to determine a dedicated relatedness threshold $\tau$ for each embedding method.

First, we constructed a dataset of word pairs, each containing either related or unrelated words. We identified the top 1,000 most common English words according to \textsc{WordNet}~\cite{wordnet}. Then, using the Merriam-Webster API, we generated 32,000 pairs of related words by finding synonyms for each common word. We also created 32,000 pairs of unrelated words by randomly pairing the original 1,000 words (ensuring no synonyms were paired together).

Second, we queried each word embedding model with each of the 64,000 pairs to obtain the semantic relatedness scores. Using a binary search, we identified the optimal threshold that balances precision and recall in discerning related and unrelated pairs for each embedding model. \tower uses these thresholds to classify each word $e_i \in \overline{E}(x)$ as "related", "unrelated", or "undefined".

\smallskip

The final output of the ensemble is derived from the relatedness scores from each model through two methods: aggregation (taking the proportion of models returning each classification) and voting (taking the classification with the most models). A weighting can also be applied to each model in the ensemble derived from their area-under-the-curve (AUC) score based on their performance on the synonym data. These are all configurations parameters of \tower.

\subsection{Trustworthiness Oracle Outcome}
Finally, \tower determines the trustworthiness of the prediction by combining the relatedness scores for each word in the filtered explanation. \tower can use several methods to combine these results: averaging (a prediction is trustworthy in proportion to the relatedness scores), plurality (a prediction is trustworthy if the majority of words in $\overline{E}(x)$ are related to $c$), and sufficiency (a prediction is trustworthy if at least one word in $\overline{E}(x)$ is related to $c$). Averaging these outcomes over many test cases provides the overall trustworthiness of the model.

\smallskip
Note that, unlike Definition 2, the oracle might have a third outcome, "undefined", due to the ensemble's uncertain responses. This approach is intended to avoid false positives, acknowledging that even humans can be indecisive when judging the plausibility of explanations~\cite{lime}.

\section{Evaluation}

We conducted a series of experiments to answer two research questions:

\begin{itemize}
    \item[\textbf{RQ1}] \textbf{Unsupervised training on noisy models}: \textit{Is it possible to train \tower configuration parameters without labelled data but using noisy data instead?}

   % Is \tower effective in detecting trustworthiness issues in models trained with noisy data?\stefano{this is a sort of unsupervised learning since we are not using labelled data. The RQ1 itself is "artificial" since the real point of RQ1 is calibrating the parameters using an unsupervised approach. I would reformulate RQ1: Is it possible to train \tower parameters without labelled data? Even better: Adding noise to data damage the semantic correlation between explanations and the labels? In the cases where the answer is yes \tower can be considered appropriate. We discover that there exist at least two cases where this condition does not apply that is "removal" and "natural" noises and thus in these conditions \tower can not work in principle}
   % \steven{In the latter case, we don't know whether it's because \tower doesn't work under those conditions as you said, or whether those noise types aren't actually a good approximation of trustworthiness. "Adding noise to data damages the semantic correlation between explanations and the labels" is the assumption we make in order to "train \tower parameters without labelled data", and also in "detecting trustworthiness issues in models trained with noisy data", so I'm unsure.}
    \item[\textbf{RQ2}] \textbf{Human-based evaluation}: \textit{How effective is the trained configuration of \tower if evaluated with human labelling of trustworthiness text classifications?}
   % \stefano{once we have proved that there exist at least two dataset that damage the semantic properties that \tower measure using embedding. After this we can proceed to measure the effectiveness of \tower in these scenarios, or prove that it doesn't work in the others.We do this test using real labelled data. In this phase we are not learning anymore so \tower remain unsupervised. Am I wrong on this point?}
   % \steven{As clarification, RQ1 shows that two types of noise damages the semantic relations, based on assumptions and artificial data. RQ2 measures the effectiveness of \tower against a ground truth, unrelated to any noise-based conditions.}
\end{itemize}

We conducted two experiments to answer these questions.

\smallskip
Datasets with explanations labelled as trustworthy or not are rare and expensive to create. Thus, for \textbf{RQ1}, we first create artificial datasets to train and evaluate the best configuration  \tower. To approximate this labelling, we make the assumption that models trained on noisy data will produce less trustworthy explanations, in proportion to the level of noise. Indeed, it is a common practice to use increasing artificial noise levels to produce relatively untrustworthy models~\cite{lime, class-noise-vs-attribute-noise, how-much-noise-is-too-much, challenges-of-text-classification-with-noisy-historical-data, comparing-automatic-and-human-evaluation-of-local-explanations}. For this, we artificially inject different levels of noise into datasets and train models on them, producing results with assumedly different proportions of trustworthy and untrustworthy explanations. We measure the relation between this increasing noise level and \tower's output. Through this, we trained the optimal tool configuration for \tower with respect to the produced untrustworthy models. Finally, we verify that we have a configuration of \tower that can detect a decrease in trustworthiness with the increase of noise level.

%\stefano{We should stress more that this is just an assumption. Indeed if we lokk at the results it is not always true. Here the idea is that we don ot want to use labelled data yet because they are rare and expensive and we proceed with an unsupervised approach. We add noise to the data ASSUMING that noise degrade trustworthiness in proportion to the noise level applied. We should make clear that this just an assumption to approximate the labelling. Indeed after a parameter calibration phase performed with noisy dataset we answer RQ2 using real labelled data }
%\steven{added}

For \textbf{RQ2}, we create a trustworthiness dataset -- that is, a dataset of explanations labelled by humans as trustworthy or not. This dataset is more in line with what would be found in real-world data than the artificial dataset -- we make no assumptions of the production of trustworthiness -- and can be considered a ground truth. We then use this data to evaluate \tower in its optimal configuration.

\subsection{Implementation}

We implemented \tool with the following five word embedding methods: \textsc{FastText}, \textsc{GloVe}, \textsc{Neural-Net Language Model (NNLM)}, \textsc{Swivel}, and \textsc{Universal Sentence Encoder (USE)}. These are popular methods commonly used in natural language processing.
The explanation technique used is \lime at 5,000 iterations~\cite{lime}.
%\steven{TODO: add versions}

\subsection{RQ1 - Unsupervised training on noisy models} \label{sec:noise-eval}

%The purpose of \tower is to detect untrustworthy models. To evaluate this, we train various models on datasets with increasing artificial noise levels to emulate relatively untrustworthy models. This is common practice in the literature~\cite{ritual, lime, class-noise-vs-attribute-noise, how-much-noise-is-too-much, challenges-of-text-classification-with-noisy-historical-data, an-effective-label-noise-model, comparing-automatic-and-human-evaluation-of-local-explanations}.
% check each to make sure it's correct

We produced untrustworthy models as follows:

%This evaluation can be broken into three steps: the training of the models, the use of \tower at various configurations, and the choosing of an analysis method. 

%First, for machine learning model training, we have the following:

\begin{itemize}

    \item \textbf{Model type.} Through scikit-learn, we chose the following four classifier models: multinomial naive Bayes (MNB), decision tree (DT), random forest (RF), and stochastic gradient descent (SGD). We picked these as they are some of the most popular machine learning algorithms. We use five-fold cross-validation to train these classifiers.
    
    \item \textbf{Dataset.} By popularity on HuggingFace\footnote{\url{https://huggingface.co/}}, we chose the following five datasets: 20 Newsgroups, AG News, DBpedia14, Emotion, and IMDB. The 20 Newsgroups dataset was also featured in the original \lime paper~\cite{lime}. Details can be found in Table~\ref{tab:datasets-eval}.
    
    \item \textbf{Noise type.} For each instance, we can create a noisy version via the following four chosen methods:
    
    \begin{itemize}
        \item \textit{Removal noise.} We remove 30\%-70\% of the words from the instance~\cite{challenges-of-text-classification-with-noisy-historical-data}.
        
        \item \textit{Label noise.} We change the label of the instance to a different one at random~\cite{class-noise-vs-attribute-noise, challenges-of-text-classification-with-noisy-historical-data}.
        
        \item \textit{Bias.} For each category in the dataset, we choose a unique random sentence\footnote{Randomly chosen from random Wikipedia articles.}. The random sentence chosen for the instance's category is appended to the end of the instance. The sentence is different between categories, but is the same sentence for each instance within the category.
        
        \item \textit{Natural noise.} This is specific for the 20 Newsgroups dataset. The original dataset has mostly-irrelevant headers and footers, which were removed. For this noise method, we re-introduce these to the instance. This was done in the experiment in the original \lime paper~\cite{lime}.
    \end{itemize}
    
\end{itemize}

\begin{table*}[]
\centering
\caption{Datasets used for evaluations.}
\label{tab:datasets-eval}
\begin{tabular}{@{}lrrlc@{}}
\toprule
\multicolumn{1}{l}{Name} & \multicolumn{1}{c}{Instances} & \multicolumn{1}{c}{Classes} & \multicolumn{1}{l}{Description} & \multicolumn{1}{c}{RQ} \\ \midrule
20 Newsgroups & 19k & 20 & A collection of newsgroup documents. & \multirow{5}{*}{RQ1\&2} \\
AG News & 128k & 4 & A collection of news articles from over 2,000 sources. &  \\
DBpedia14 & 630k & 14 & A collection of texts from DBpedia 2014. &  \\
Emotion & 20k & 6 & A collection of English Twitter messages. &  \\
IMDb & 50k & 2 & A collection of movie reviews from IMDb. &  \\ \midrule
Amazon Review Polarity & 4,000K & 2 & A collection of Amazon reviews across 18 years. & \multirow{3}{*}{RQ2} \\
Rotten Tomatoes & 11k & 2 & A collection of movie reviews from Rotten Tomatoes. &  \\
Yahoo Answers Topics & 2,000K & 10 & A collection of texts from Yahoo! Answers. &  \\ \bottomrule
\end{tabular}
\end{table*}

\smallskip
We do this for five levels of noise: 0\%, 25\%, 50\%, 75\%, and 100\%. These levels indicate what percentage of instances of the training set are injected with noise. At 25\%, for example, 25\% of the training instances are randomly chosen and replaced with their noisy versions.

\smallskip
For each noise type, the test set is the same for each noise level. It consists of all instances from the original clean test set combined with the noisy version of each instance, doubling the set's overall size.
We do this as we want to see the models' explanations for both clean and noisy data. We also want all models of each noise type to be tested on the same data, regardless of their noise levels.
% I feel like we should have used just the clean version to test, given the golden rule of machine learning is to not change the test data
This results in a total of 64 sets of models being trained (4 models $\times$ 5 datasets $\times$ 3 noise types + 4 models (natural noise only for 20 Newsgroups) = 64), with 25 models in each set; five-fold for each of the five noise levels (totaling around 1,600 models). The 0\% noise models are the same model across each noise type. %\stefano{I ma sure is correct but I can not figure out easily where this 64 is coming from...  are  testing models just normal models? In this case delete the "testing"} \steven{4 models x 5 datasets x 3 noise types + 4 models (natural noise) = 64. Should I include that? "testing" deleted.}

\smallskip
We then have another set of parameters for the tool itself that we aim to learn with unsupervised learning using the noisy models. These are:

\begin{itemize}
    \item \textbf{Word embedding exclusion range.} \{\textit{0, 0.07}\} This controls the range around the thresholds of the word embedding methods for which the relation between two words is considered \textit{undefined}.
    
    \item \textbf{Word embedding weighting.} \{\textit{true, false}\} This determines whether the word embeddings in the ensemble should be weighted by their AUC scores.
    
    \item \textbf{Relatedness score calculation method.} \{\textit{aggregation, voting}\} This determines how the word embedding outputs are combined in the ensemble. \textit{Aggregation} is the percentage of methods that outputted each of \textit{related}, \textit{unrelated}, and \textit{undefined}. \textit{Voting} is the simple plurality of results.
    
    \item \textbf{Explanation threshold.} \{\textit{0, 0.05}\} Only explanatory words with importance scores above this threshold are considered.
    
    \item \textbf{Explanation top-\textit{n}.} \{\textit{5, 10}\} Only the top \textit{n} words of the explanation are considered.
    
    \item \textbf{Trustworthiness score calculation method.} \{\textit{average, plurality, sufficiency}\} This is how the trustworthiness of a prediction is calculated from the relatedness of words. 
    
    \begin{itemize}
        \item \textit{Average} is the average of each of \textit{related}, \textit{unrelated}, and \textit{undefined}, considered as \textit{trustworthy}, \textit{untrustworthy}, and \textit{undefined}. For example, an explanation of 2 words with relatedness scores of $(0.0,0.8,0.2)$ and $(1.0,0.0,0.0)$ would have a trustworthiness score of $(0.5,0.4,0.1)$.
       % \stefano{this average is not clear fro me. Average among cosine similarity score of the words in explanation to rate the overall model trustworthiness. All the other mesure rate the single prediction. Please put this more in evidence. Single prediction trustworthiness versus average model trustworthiness } \steven{clarified}

        \item \textit{Plurality} simply takes the largest value of the three.
        \item \textit{Sufficiency} considers the prediction trustworthy if any of the words is related. This last option is considered as research has shown that humans may only require one reason to trust something \cite{explanation-insights-from-the-social-sciences}.
    \end{itemize}
     
\end{itemize}

This results in 96 possible tool configurations.
\smallskip
By running \tower using every combination of both the model and tool configurations, we produce a resultant trustworthiness score for each model. This is a 3-tuple in the form of (\textit{trustworthiness}, \textit{untrustworthiness}, \textit{undefined}), summing to one.

\smallskip
% better lead-in
We now wish to find the optimal configuration for \tower; that is, the configuration with the greatest monotonic decrease in predicted trustworthiness with increasing noise level.
%\valerio{find best configuration that have least difference between these slope types, EXMPLAIN HOW} \steven{now explained below}

\smallskip
Here, several questions arise. Firstly, obtaining the slope of noise level against trustworthiness score requires a single value for each data point; yet, we have three values for each. Next, the fact that increasing the noise level likely decreases accuracy, which may affect results. Finally, it is possible that the noise types will not be equally good at representing untrustworthiness. Thus, we have another set of options for the analysis of these results:

\begin{itemize}
    \item \textbf{Slope calculation method.} We can calculate the slope either by taking only the decrease in \textit{trustworthiness}; taking only the increase in \textit{untrustworthiness}; or taking the decrease in $\textit{trustworthiness} / (\textit{trustworthiness} + \textit{untrustworthiness})$.
    \item \textbf{Adjusted slope.} Currently, for each noise level, all correctly-predicted instances are considered for the slope calculation, regardless of if they are predicted incorrectly by models at other noise levels. To reduce the impact of decreasing accuracy with increasing noise level, the calculation can be adjusted such that, for every noise level, it discards any instance that is not predicted correctly across all noise levels (following Definition~1).
    %\stefano{is this coming from the definition of trustworthiness itself? I am not sure this is a configurable parameter} \steven{have attempted to clarify}
    \item \textbf{Noise types.} What combination of noise types to consider.
\end{itemize}

\smallskip
This results in 48 analysis methods to determine the effectiveness of a tool configuration (or 96 for the 20 Newsgroups dataset).

\smallskip
We have various possible configurations for model training, tool running, and analysis. We use these now to find the optimal tool configuration. 

\begin{figure}[t!]
    \centering
    \begin{subfigure}[ht]{\linewidth}
        \centering
        \includegraphics[width=\linewidth]{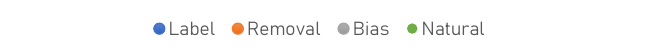}
    \end{subfigure}
    \begin{subfigure}[ht]{0.49\linewidth}
        \centering
        \includegraphics[width=\linewidth]{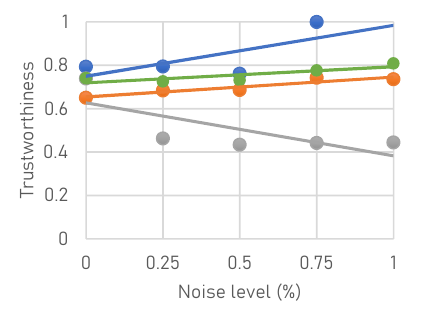}
        \caption{multinomial naive Bayes}
    \end{subfigure}%
    \hfill
    \begin{subfigure}[ht]{0.49\linewidth}
        \centering
        \includegraphics[width=\linewidth]{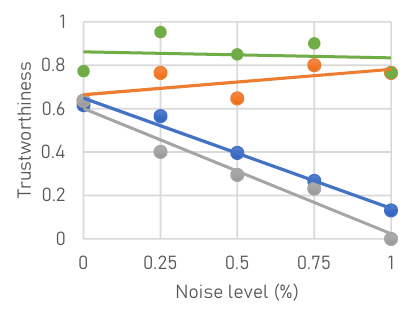}
        \caption{decision tree}
    \end{subfigure}
    \hfill
    \begin{subfigure}[ht]{0.49\linewidth}
        \centering
        \includegraphics[width=\linewidth]{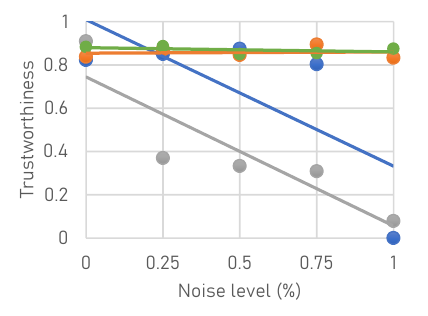}
        \caption{random forest}
    \end{subfigure}
    \hfill
    \begin{subfigure}[ht]{0.49\linewidth}
        \centering
        \includegraphics[width=\linewidth]{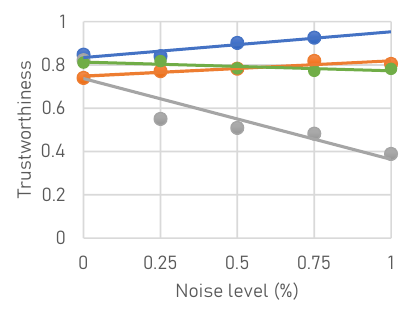}
        \caption{stochastic gradient decent}
    \end{subfigure}
    \caption{Example charts of trustworthiness against noise level for each model using optimal configuration and $t/(t+u)$ slope calculation for the 20 Newsgroup dataset. Adjusted slope used for all except label noise.}
    \label{fig:noise-charts}
\end{figure}

\smallskip
For each tool configuration, we run \tower on a chosen model set. This produces a set of five trustworthiness scores, one for each noise level. We wish to find the variation of scores across the noise levels. To model this, we fit a line to these points and suppose that the more negative the slope, the greater the decrease of trustworthiness. This slope is calculated based on a chosen analysis method. An example of these slopes can be seen in Fig.~\ref{fig:noise-charts}.

\smallskip
For each model set, the tool configurations are ranked based on their produced slope. These rankings are summed across all model sets to produce an overall score for each tool configuration. Rankings are used as the slopes cannot be directly compared between the various different options.

\smallskip
This is done for each analysis method. To reduce any bias between the slope calculation types, the analysis method with the least rank variation between these types is selected. This is as we consider each slope calculation type as an equally valid way to measure the slope.

\smallskip
Finally, the tool configuration with the highest overall score for that analysis method is chosen as the most optimal.

%\stefano{The idea is that  we want to see as much variation as possible in trustworthiness scores across different noise levels monotonically descending at increasing noise levels. We decide to model this decreasing scores fitting a line. The greater the linear coefficient the largest the trustworthiness variation. Explain better this assumption. } \stefano{I am not sure what is the rationale behind this further assumption. I have some intuition but is a bit vague. Anyway we need to make this explicit!} \steven{added}

\medskip
\textbf{RQ1 Results} We evaluated \tower on its \textit{effectiveness on noisy models} (\textbf{RQ1}) and found that, assuming increasing noise in the training data of models decreases its trustworthiness, it does detect some form of trustworthiness. Table~\ref{tab:noise-table} shows the slopes averaged across all slope calculation methods of trustworthiness scores against noise level for \tower at its optimal configuration. Adjusted slopes are used for non-label noise. The more negative the slope, the more the trustworthiness score decreased with increasing noise percentage. 

As an example, the \textit{decision tree} models trained on the \textit{20 Newsgroups} dataset using \textit{bias} has an adjusted slope of $-0.428$ using the $\textit{trustworthiness} / (\textit{trustworthiness} + \textit{untrustworthiness})$ slope calculation. This indicates an overall decrease of $0.428$ in trustworthiness between the models at 0\% and 100\% noise. In contrast, the same using \textit{removal noise} has a positive slope of $0.220$, indicating an unusual increase of trustworthiness score with increasing noise.

We see that, between all model types, this correlation occurs strongly across label and bias noise and moderately across natural noise. 

We also see that \tower did not find much difference in the use of removal noise. This indeed makes sense, as the removal of words in the training set of the models would likely decrease model accuracy, but should not overmuch alter the explanation of correct predictions -- especially with the slope calculation adjusted to only consider predictions correct across all noise levels. The semantic correlation that \tower uses should remain consistent. This is evidence towards the procedure of this evaluation not introducing unwanted biases into \tower itself.
%\steven{does this make sense? should this be moved to discussion?}

%\stefano{This is an expected results since \tower is measuring the semantic correlations among labels and explanations. Just removing words is likely to decrease the models accuracy but for correct guess it does not alter in principle the semantic correlation. And we are adjusting score for considering correctness variations indeed. This prove that the unsupervised procedure followed for answering RQ1 is fundamentally correct and it is not introducing spurious correlations or other biases despite the complex procedure and the numerous parameters. It is a passed sanity check. Congratulations it is a good news sell this point as a success because it is.} \steven{added}

%\valerio{explain what the values mean in the table, give example explain the outlier }
%\steven{TODO: results summary}

Of note, an error resulted in the loss of one set of data, which is absent in the Table (the \textit{random forest} models trained on the \textit{DBpedia14} dataset using \textit{label noise}).

\begin{table}[]
\caption{Slopes for various datasets, model types, and noise types using optimal configuration and $t/(t+u)$ slope calculation. Adjusted slopes used for non-label noise.}
\label{tab:noise-table}
\begin{tabular}{@{}ll|cccc@{}}
\toprule
\multicolumn{2}{c|}{Test models} & \multicolumn{4}{c}{Noise types} \\ \midrule
\multicolumn{1}{c}{Dataset} & \multicolumn{1}{c|}{Model} & \multicolumn{1}{c}{Bias} & \multicolumn{1}{c}{Label} & \multicolumn{1}{c}{Removal} & \multicolumn{1}{c}{Natural} \\ \midrule
 & MNB & \cellcolor[HTML]{E5E57E}-0.100 & \cellcolor[HTML]{D2E07A}-0.176 & \cellcolor[HTML]{FCEA83}-0.010 & \cellcolor[HTML]{FFE983}0.012 \\
 & DT & \cellcolor[HTML]{91D16D}-0.428 & \cellcolor[HTML]{E0E47D}-0.118 & \cellcolor[HTML]{FFB867}0.220 & \cellcolor[HTML]{DEE37D}-0.129 \\
 & RF & \cellcolor[HTML]{87CF6B}-0.467 & \cellcolor[HTML]{C7DE78}-0.220 & \cellcolor[HTML]{FBEA83}-0.014 & \cellcolor[HTML]{E5E57E}-0.101 \\
\multirow{-4}{*}{20 Newsgroups} & SGD & \cellcolor[HTML]{C0DC77}-0.246 & \cellcolor[HTML]{C9DE79}-0.211 & \cellcolor[HTML]{F3E881}-0.044 & \cellcolor[HTML]{F2E881}-0.050 \\ \midrule
 & MNB & \cellcolor[HTML]{E5E57E}-0.101 & \cellcolor[HTML]{E3E47E}-0.109 & \cellcolor[HTML]{FFEB84}0.003 & - \\
 & DT & \cellcolor[HTML]{99D36F}-0.400 & \cellcolor[HTML]{D5E17B}-0.162 & \cellcolor[HTML]{FFDF7E}0.052 & - \\
 & RF & \cellcolor[HTML]{BDDB76}-0.257 & \cellcolor[HTML]{D4E17B}-0.168 & \cellcolor[HTML]{F8E982}-0.025 & - \\
\multirow{-4}{*}{AG News} & SGD & \cellcolor[HTML]{AED873}-0.315 & \cellcolor[HTML]{B8DA75}-0.275 & \cellcolor[HTML]{FFE882}0.016 & - \\ \midrule
 & MNB & \cellcolor[HTML]{D5E17B}-0.162 & \cellcolor[HTML]{9BD36F}-0.390 & \cellcolor[HTML]{FFE681}0.023 & - \\
 & DT & \cellcolor[HTML]{91D16D}-0.430 & \cellcolor[HTML]{A8D672}-0.340 & \cellcolor[HTML]{F8E982}-0.024 & - \\
 & RF & \cellcolor[HTML]{6EC966}-0.567 &  & \cellcolor[HTML]{F8E982}-0.025 & - \\
\multirow{-4}{*}{DBpedia14} & SGD & \cellcolor[HTML]{C7DE78}-0.219 & \cellcolor[HTML]{92D16D}-0.425 & \cellcolor[HTML]{FFE581}0.029 & - \\ \midrule
 & MNB & \cellcolor[HTML]{C9DE79}-0.210 & \cellcolor[HTML]{C5DD78}-0.227 & \cellcolor[HTML]{FCEA83}-0.009 & - \\
 & DT & \cellcolor[HTML]{A0D570}-0.369 & \cellcolor[HTML]{C9DE79}-0.211 & \cellcolor[HTML]{FFD97A}0.078 & - \\
 & RF & \cellcolor[HTML]{7BCC69}-0.514 & \cellcolor[HTML]{ADD873}-0.320 & \cellcolor[HTML]{FCEA83}-0.008 & - \\
\multirow{-4}{*}{Emotion} & SGD & \cellcolor[HTML]{90D16D}-0.433 & \cellcolor[HTML]{B2D974}-0.302 & \cellcolor[HTML]{FCEA83}-0.012 & - \\ \midrule
 & MNB & \cellcolor[HTML]{EDE680}-0.070 & \cellcolor[HTML]{F7E982}-0.030 & \cellcolor[HTML]{F5E882}-0.037 & - \\
 & DT & \cellcolor[HTML]{A9D772}-0.334 & \cellcolor[HTML]{F1E781}-0.052 & \cellcolor[HTML]{FFE682}0.022 & - \\
 & RF & \cellcolor[HTML]{C3DD77}-0.233 & \cellcolor[HTML]{F1E781}-0.051 & \cellcolor[HTML]{FFD075}0.117 & - \\
\multirow{-4}{*}{IMDb} & SGD & \cellcolor[HTML]{ADD873}-0.321 & \cellcolor[HTML]{EBE680}-0.077 & \cellcolor[HTML]{FFE883}0.014 & - \\ \bottomrule
\end{tabular}
\end{table}

In addition, we found that the analysis methods with the least rank variation between the slope calculation types uses adjusted slope, as well as using only bias as the noise type. Thus, of the noise injection methods, bias seems to be the most representative of creating untrustworthiness.
% isn't the option of choosing which combination of noise types is best just overfitting??

\smallskip
To note, we find that label noise does not evaluate correctly with the adjusted slope, as the result of having 100\% label noise is 0\% accuracy. This would result in zero instances considered while using the adjusted method. Thus, any analysis method including both adjusted slope and label noise is not considered. 
%\stefano{This sentence looks correct to me. I don't understand where is the problem. 100 noise means very low accuracy. No ? I would rather observe that label noise has very high trustworthiness scores in some picture in Figure 2. Why ? Sorry but it's not clear why.} \steven{clarified}

% Using this analysis method, we find that the best configuration combination is: word_embedding_precision_modifier – 1.0, word_embedding_exclusion_range – 0.07, do_word_embedding_weighting – false, calculate_word_scores_method – plurality, explanation_threshold – 0.05, explanation_top_n – 10, calculate_instance_scores_method – plurality. We can verify that, respectively for the three slope calculation types, this configuration is ranked 18, 8 and 1 out of a maximum of 96. 

% Interestingly, we also see that, for the 20 Newsgroups dataset, removal and natural noise have a similar slope. However, both are shallow, indicating little correlation to trustworthiness, despite what is shown in the \lime paper~\cite{lime}.

\subsection{RQ2 - Human-based evaluation}

We wish to know whether \tower's trustworthiness predictions align with humans' perception of trustworthiness. To do this, we obtain explanations from various models, using human judgement to determine their trustworthiness, creating a ground truth to compare to. Three authors of the paper performed the labelling.

\smallskip
The explanations to label are generated through pretrained models fine-tuned on various datasets, chosen by popularity on HuggingFace. These are: 20 Newsgroups, AG News, Amazon Review Polarity, DBpedia14, Emotion, IMDb, Rotten Tomatoes, and Yahoo Answers Topics. Details can be found in Table~\ref{tab:datasets-eval}.
%\steven{TODO: add links for models and datasets}
% \steven{should we have a table for these as well?}
We chose to use pretrained models as they more closely align with real world application than models trained ourselves. This allows us to test on more realistic instances of trustworthy and untrustworthy explanations. We assume the generalisability of \tower since it evaluates the explanations independently from the model that generates the predictions.
%\stefano{We assume this is possible since \tower learn just to evaluate the semantic correlations between words in explanations and labels independently from the model that generated the predictions.} \steven{added}

From each model and dataset, we randomly choose 1,000 instances from each dataset and explanations generated using \lime. We then run \tower, using the best configuration trained in Section~\ref{sec:noise-eval}, on these explanations. Of the 8,000 instances, 501 are randomly chosen with a proportion of $33\%\pm7.5\%$ between ones predicted as \textit{trustworthy}, \textit{untrustworthy}, and \textit{undefined}. This is done to combat the likely trustworthy skew in the data. The error of $\pm7.5\%$ is added to prevent potential bias from knowing the exact ratio of explanations.

The labelling procedure is thus: Each person is given an instance and its possible classes. They must predict the class. This is to ensure that they consider the text and not simply give a label based on only the explanation~\cite{whose-right-reasons}.
% double check this citation
Next, the explanation is given -- the top ten words, highlighting the respective words in the text as well as giving the importance scores. They may choose to label this explanation as trustworthy, untrustworthy or undefined. If the person does not predict the correct class, the instance is discarded; this is not shown to the human.

We have two label each instance. If two authors do not agree on the label, the third author is brought in. If none agree, the instance is discarded. These labelled instances are then compared to the predictions from \tower to evaluate its effectiveness.

Of the 501 instances, 139 were discarded for a participant incorrectly predicting the category and 15 for a lack of agreement of trustworthiness label. Another 19 were erroneously discarded. This resulted in a final set of 328 instances.

\medskip
\textbf{RQ2 Results} We evaluated \tower against \textit{human-based evaluation} (\textbf{RQ2}). To do so, we created our own dataset of human-labelled ground truths for trustworthy and untrustworthy explanations. In total, 501 instances were labelled, 328 of which were considered. As seen in Table~\ref{tab:human-labels}, 293 (89\%) were labelled \textit{trustworthy}, 21 (6\%) \textit{untrustworthy}, and 14 (4\%) \textit{undefined}.

\begin{table}[]
\caption{Human-labelled data.}
\label{tab:human-labels}
\begin{tabular}{@{}lrrr|r@{}}
\toprule
Dataset & Untrust. & Trust. & Undefined & Total \\ \midrule
20 Newsgroups & 1 & 25 & 0 & 26 \\
AG News & 3 & 41 & 2 & 46 \\
Amazon Reviews Polarity & 7 & 42 & 2 & 51 \\
DBpedia14 & 1 & 42 & 0 & 43 \\
Emotion & 0 & 27 & 0 & 27 \\
IMDb & 1 & 45 & 3 & 49 \\
Rotten Tomatoes & 6 & 39 & 6 & 51 \\
Yahoo Answers Topics & 2 & 32 & 1 & 35 \\ \midrule
Total & 21 & 293 & 14 & 328 \\ \bottomrule
\end{tabular}
\vspace{-2mm}
\end{table}

\begin{figure}[t!]
    \centerline{\includegraphics[width=0.5\textwidth]{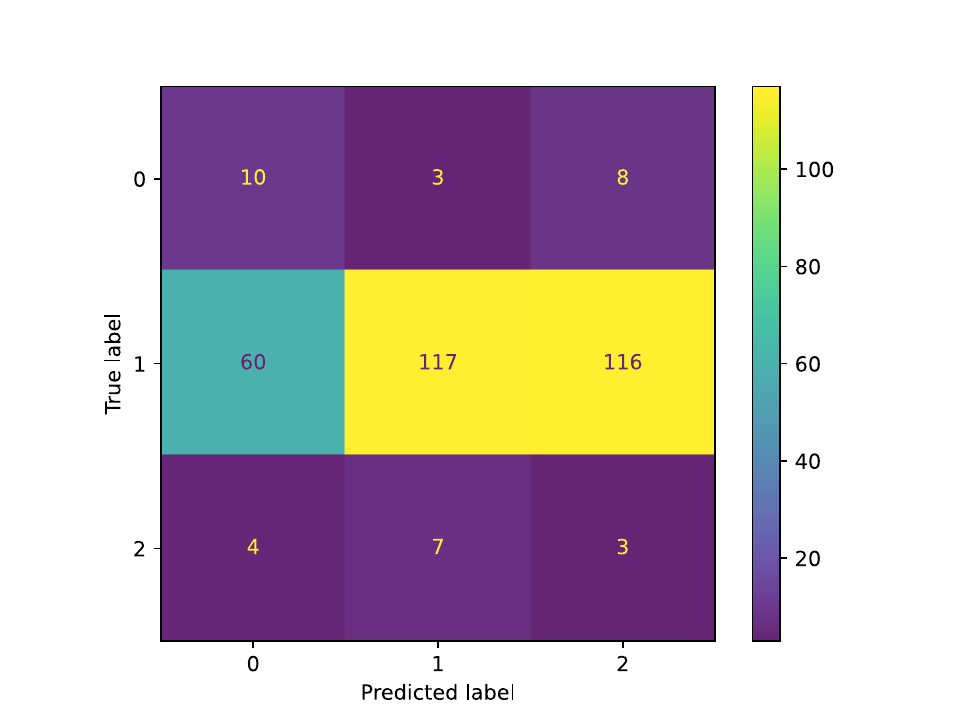}}
    \caption{Confusion matrix. Labels 0, 1, and 2 are trustworthy, untrustworthy, and undefined, respectively.}
    \label{fig:eval1-matrix}
\end{figure}

Fig.~\ref{fig:eval1-matrix} shows the confusion matrix of the optimal configuration of \tower on the human-labelled ground truth. The precision, recall, and F1 score for labelling as trustworthy is 0.92, 0.40, and 0.56, respectively. For labelling as untrustworthy, these are 0.14, 0.48, and 0.21.
% If we disregard undefined, these go to 0.98, 0.66, 0.76 and 0.14, 0.77, 0.22, respectively.

\subsection{Discussion}

From our noise-based evaluation (\textbf{RQ1}), we find that there is generally a correlation between \tower's output and the trustworthiness of a model, supposing that artificially induced noise into the training data of a model is a good approximation of untrustworthiness. 
% \stefano{add here my consideration on the semantic correlation from the previous subsection. There are justified exception. and in the conclusion I would remember that this trustworthiness that we measure is a semantic correlation} 
We find that bias is better at this approximation than label or removal noise, assuming that \tower's predictions are accurate. We also see from Table~\ref{tab:noise-table} that \tower is not able to detect models trained on removal noise very well. This may be because of our specific implementation of the noise, or because machine learning models are often robust to this type of noise \cite{how-much-noise-is-too-much}.

From our human-based evaluation (\textbf{RQ2}), however, we find that the introduction of noise may not be a good approximation for untrustworthiness after all.
\tower, using the optimal configuration found from the noise-based evaluation, performed poorly in against human-labelled ground truths of trustworthy explanations. We find that it has a bias against predicting as \textit{trustworthy}.
% It has a bias against predicting 'trustworthy'?
% If it predicts 'untrustworthy', it is more likely that it is 'trustworthy' (low 'untrustworthy' precision)
Noise can introduce spurious correlations that can be detected via a semantic measure like \tower. % Keep this?
It is also possible that trustworthy explanations found in our simpler models may be different to that of pretrained models.
%\stefano{noise can introduce spurious correlations. if the model use this correlation for prediction then a smatic measure can discover this situation. If the noise, like in the removal case, does not induce spurious correlations then \tower can not work. The noise itself does not approximate trustworthiness.} \steven{added}

We can gain insight by looking at the incorrectly-labelled instances. Of the 328 ground-truth instances, \tower labels 198 incorrectly. We are more interested in the \textit{trustworthy} and \textit{untrustworthy} instances, so we will disregard for now the 135 instances labelled as \textit{undefined} by either \tower or humans. Of the 63 remaining, 60 are falsely labelled \textit{trustworthy}. Of the 63, 32 are from the Amazon Polarity, Rotten Tomatoes, and IMDb datasets; these all have the categories 'positive' and 'negative'. For the rest, the most common categories are 'joy', 'sports', 'science', 'sadness', and 'health'. From this, we can intuit that the effectiveness of \tower is somewhat proportional to how specific the class name is. The more general the class, the harder it is to relate words to it.

This seems to be, in part, due to our limited method of measuring trustworthiness. \tower only measures the relatedness between the explanation words and the singular class name. To alleviate this, a possible solution is to have the category instead be a collection of words describing the concept in more detail, such as a definition.

Other possible areas of improvement include taking into account words in context (e.g. "not good"), importance scores, unusually impactful singular words, other class names, incorrect model predictions, and the relationship between words in the explanation.

Finally, a major way forward would be to curate a larger and less-skewed dataset so to more effectively evaluate the tool.

%BIAS real world trustworthiness, noise might not reflect 

\section{Threats to Validity}

\textit{Generalisability of results.} One threat of external validity is that the evaluation results dot not generalise for other models and dataset. We mitigate this threat by using various different datasets and models in the evaluation. The noise-based evaluation uses 5 datasets and 4 models, while the human-based evaluation used 8 datasets and models. This reduces the chance of individual datasets or models biasing the results.

\textit{Bias of word embedding methods.} Word embeddings may have their own bias, errors, and trustworthiness issues. Are we not simply moving trust from the machine learning model to these embedding methods? To alleviate this, an ensemble of methods is used to reduce the impact of errors in any single embedding method.

\textit{Human collection of ground truth.} Humans may have biases and failings when labelling explanations as trustworthy or not. To alleviate this, two authors of this paper did the labelling independently, and final labels were picked only with majority vote involving a third independent author. The participants were also made to choose a class for the input first, incentivising them to consider the input in its entirety when labelling. In addition, the proportion of potentially trustworthy and untrustworthy instances was somewhat randomised, preventing bias towards labelling for an equal ratio.
As the human labellers knew how \tower worked, they were instructed to label the data using only their human intuition, disregarding any knowledge of the tool.

%\valerio{the humna labeller knew how \tower works }
%\steven{is this a good enough reason?}

% Faithfulness of explanations. The thought of \lime being incorrect has not been considered at all.

\section{Related Work}

Trustworthiness AI and ML is an important research topic that has gained a lot of attention~\cite{floridi2021establishing}. To the best of our knowledge, there are no prior attempts to create an automated trustworthiness oracle. \tower is the first automated trustworthiness technique. 

\smallskip
There have been metrics proposed for measuring trustworthiness in machine learning models. Schmidt and Biessmann~\cite{quantifying-interpretability-and-trust} proposes a metric based on the increase in information transfer rate for humans doing a task – that is, the change in speed and accuracy of human interpreters when assisted by a machine explanation or not. The Trustworthy Explainability Acceptance metric requires industry experts to evaluate the system in its calculations~\cite{Trustworthy-Explainability-Acceptance-Medical, Trustworthy-Explainability-Acceptance-Food}. There is the building of trust probabilities based on subjective logic via uncertainty given by a human observer~\cite{there-is-hope-after-all}. RITUAL is a collection of metrics that require humans to manually evaluate~\cite{ritual}. All of these require humans in the loop, however; \tower aims to evaluate trustworthiness automatically.

\smallskip
The use of explainability in ML testing has also been explored. There have been testing on the effectiveness of individual explanatory techniques~\cite{evaluating-the-quality-of-machine-learning-explanations} and testing of the evaluation of explanations~\cite{quantitative-evaluation-of-machine-learning-explanations, explaining-explanations-an-overview-of-interpretability}. There has been research to mitigate potential unreliability in the explanations themselves~\cite{the-unreliability-of-saliency-methods, explanations-can-be-manipulated}. More relevant to our goal, there have been research on evaluating models based on explanations~\cite{quantifying-interpretability-and-trust}, as mentioned previously, including the original \lime paper itself~\cite{lime}. These works also require humans in the loop.

\smallskip
In addition, the current work towards the measurement of plausibility of explanations seems to lie mainly within the natural language processing field. They involve procedures such as comparing against human-created ground truths~\cite{eraser, invariant-rationalization, interpretable-neural-predictions-with-differentiable-binary-variables, automatic-metrics-for-the-evaluation-of-natural-language-explanations}, or asking people directly~\cite{learning-to-faithfully-rationalize-by-construction}. Again, we have found no fully automated way to do this.

\smallskip
Other related works are techniques on automated fairness testing.  They do not rely on word embeddings but use \lime explanations to check whether the predictions were influenced by the presence of sensitive or protected attributes (e.g., gender, ethnicity, age). For example, Aggarwal et al.~\cite{black-box-fairness-testing} proposes an automated method for the evaluation of fairness, via the use of the local linear model created by \lime. However, fairness by itself does not constitute a full evaluation of trustworthiness; our technique focuses on evaluating the justification behind the explanations.

\smallskip
The most related work of \tower is that of Jiang et al.~\cite{to-trust-or-not-to-trust-a-classifier}, which defines a 'trust score'; a measure of the level of agreement between a classifier and a modified nearest-neighbor classifier for a single response. This is a rather different approach to our proposed technique as it only rely on the agreement between multiple classifiers to measure trustworthiness. Instead, \tower analyses  the decision-making process of ML classifiers to measure trustworthiness. 

%, as it relies on processing the test data using a distance metric, its effectiveness depending on the number of dimensions.

\section{Conclusion}
% Lam: One way to achieve this is to go beyond classification correctness. We should make sure that the rationale behind correct predictions is meaningful, and thus the ML model will likely generalize well to unseen data. This is the concept of trustworthiness in ML models. 

% Stefano: "and in the conclusion I would remember that this trustworthiness that we measure is a semantic correlation"
% Stefano: "semantic relatedness is not enough for trustworthiness"

This paper presented \tower, one of the first attempts to automatically perform trustworthiness testing. The experiments with noisy data suggests that the idea behind \tower is indeed relevant in identifying trustworthiness issues. However, the poor results on the human labelled data hint that semantic unrelatedness between explanations and class names might not characterise all types of 
untrustworthy predictions.

%be the only reason for untrustworthy predictions.

%\tower. We created two trustworthiness datasets to test \tower. We find that it does not do so well.

This paper sparks exciting future work in this new research area of automated trustworthiness testing. More research is needed to understand the relationship between explanations and trustworthiness issues. 

%Future work: improving relatedness calculation, improving dataset, investigating what it means for someone to trust an explanation, extension to images

% final checklist:
% check tense

\bibliography{bibliography}{}

\begin{thebibliography}{10}

\bibitem{how-much-noise-is-too-much}
Sumeet Agarwal, Shantanu Godbole, Diwakar Punjani, and Shourya Roy.
\newblock How much noise is too much: A study in automatic text classification.
\newblock In {\em Seventh IEEE International Conference on Data Mining (ICDM 2007)}, pages 3--12, 2007.

\bibitem{black-box-fairness-testing}
Aniya Aggarwal, Pranay~Kr. Lohia, Seema Nagar, Kuntal Dey, and Diptikalyan Saha.
\newblock Black box fairness testing of machine learning models.
\newblock {\em Proceedings of the 2019 27th ACM Joint Meeting on European Software Engineering Conference and Symposium on the Foundations of Software Engineering}, 2019.

\bibitem{interpretable-neural-predictions-with-differentiable-binary-variables}
Jasmijn Bastings, Wilker Aziz, and Ivan Titov.
\newblock Interpretable neural predictions with differentiable binary variables.
\newblock In Anna Korhonen, David Traum, and Llu{\'\i}s M{\`a}rquez, editors, {\em Proceedings of the 57th Annual Meeting of the Association for Computational Linguistics}, pages 2963--2977, Florence, Italy, July 2019. Association for Computational Linguistics.

\bibitem{typology-of-the-concept-of-explanation}
Federico Cabitza, Andrea Campagner, Gianclaudio Malgieri, Chiara Natali, David Schneeberger, Karl Stoeger, and Andreas Holzinger.
\newblock Quod erat demonstrandum? - towards a typology of the concept of explanation for the design of explainable ai.
\newblock {\em Expert Syst. Appl.}, 213(PA), mar 2023.

\bibitem{ritual}
Alberto~Huertas Celdrán, Jan Bauer, Melike Demirci, Joel Leupp, Muriel~Figueredo Franco, Pedro~M. Sánchez~Sánchez, Gérôme Bovet, Gregorio~Martínez Pérez, and Burkhard Stiller.
\newblock Ritual: a platform quantifying the trustworthiness of supervised machine learning.
\newblock In {\em 2022 18th International Conference on Network and Service Management (CNSM)}, pages 364--366, 2022.

\bibitem{invariant-rationalization}
Shiyu Chang, Yang Zhang, Mo~Yu, and Tommi Jaakkola.
\newblock Invariant rationalization.
\newblock In Hal~Daumé III and Aarti Singh, editors, {\em Proceedings of the 37th International Conference on Machine Learning}, volume 119 of {\em Proceedings of Machine Learning Research}, pages 1448--1458. PMLR, 13--18 Jul 2020.

\bibitem{there-is-hope-after-all}
Mingxi Cheng, Shahin Nazarian, and Paul Bogdan.
\newblock There is hope after all: Quantifying opinion and trustworthiness in neural networks.
\newblock {\em Frontiers in artificial intelligence}, 3:54, 2020.

\bibitem{automatic-metrics-for-the-evaluation-of-natural-language-explanations}
Miruna-Adriana Clinciu, Arash Eshghi, and Helen Hastie.
\newblock A study of automatic metrics for the evaluation of natural language explanations.
\newblock In Paola Merlo, Jorg Tiedemann, and Reut Tsarfaty, editors, {\em Proceedings of the 16th Conference of the European Chapter of the Association for Computational Linguistics: Main Volume}, pages 2376--2387, Online, April 2021. Association for Computational Linguistics.

\bibitem{eraser}
Jay DeYoung, Sarthak Jain, Nazneen~Fatema Rajani, Eric Lehman, Caiming Xiong, Richard Socher, and Byron~C. Wallace.
\newblock {ERASER}: {A} benchmark to evaluate rationalized {NLP} models.
\newblock In Dan Jurafsky, Joyce Chai, Natalie Schluter, and Joel Tetreault, editors, {\em Proceedings of the 58th Annual Meeting of the Association for Computational Linguistics}, pages 4443--4458, Online, July 2020. Association for Computational Linguistics.

\bibitem{explanations-can-be-manipulated}
Ann-Kathrin Dombrowski, Maximilian Alber, Christopher~J. Anders, Marcel Ackermann, Klaus-Robert M\"{u}ller, and Pan Kessel.
\newblock {\em Explanations can be manipulated and geometry is to blame}.
\newblock Curran Associates Inc., Red Hook, NY, USA, 2019.

\bibitem{floridi2021establishing}
Luciano Floridi.
\newblock Establishing the rules for building trustworthy ai.
\newblock {\em Ethics, Governance, and Policies in Artificial Intelligence}, pages 41--45, 2021.

\bibitem{explaining-explanations-an-overview-of-interpretability}
Leilani~H. Gilpin, David Bau, Ben~Z. Yuan, Ayesha Bajwa, Michael Specter, and Lalana Kagal.
\newblock Explaining explanations: An overview of interpretability of machine learning.
\newblock In {\em 2018 IEEE 5th International Conference on Data Science and Advanced Analytics (DSAA)}, pages 80--89, 2018.

\bibitem{learning-to-faithfully-rationalize-by-construction}
Sarthak Jain, Sarah Wiegreffe, Yuval Pinter, and Byron~C. Wallace.
\newblock Learning to faithfully rationalize by construction.
\newblock In {\em Annual Meeting of the Association for Computational Linguistics}, 2020.

\bibitem{to-trust-or-not-to-trust-a-classifier}
Heinrich Jiang, Been Kim, Melody~Y. Guan, and Maya Gupta.
\newblock To trust or not to trust a classifier.
\newblock In {\em Proceedings of the 32nd International Conference on Neural Information Processing Systems}, NIPS'18, page 5546–5557, Red Hook, NY, USA, 2018. Curran Associates Inc.

\bibitem{Trustworthy-Explainability-Acceptance-Medical}
Davinder Kaur, Suleyman Uslu, Arjan Durresi, Sunil~V. Badve, and Murat Dundar.
\newblock Trustworthy explainability acceptance: A new metric to measure the trustworthiness of interpretable ai medical diagnostic systems.
\newblock In {\em Computational Intelligence in Security for Information Systems}, 2021.

\bibitem{the-unreliability-of-saliency-methods}
Pieter-Jan Kindermans, Sara Hooker, Julius Adebayo, Maximilian Alber, Kristof~T. Sch{\"u}tt, Sven D{\"a}hne, D.~Erhan, and Been Kim.
\newblock The (un)reliability of saliency methods.
\newblock In {\em Explainable AI}, 2017.

\bibitem{challenges-of-text-classification-with-noisy-historical-data}
R.~Andrew Kreek and Emilia Apostolova.
\newblock Training and prediction data discrepancies: Challenges of text classification with noisy, historical data.
\newblock In Wei Xu, Alan Ritter, Tim Baldwin, and Afshin Rahimi, editors, {\em Proceedings of the 2018 {EMNLP} Workshop W-{NUT}: The 4th Workshop on Noisy User-generated Text}, pages 104--109, Brussels, Belgium, November 2018. Association for Computational Linguistics.

\bibitem{On-the-Relation-of-Trust-and-Explainability}
Lena Kästner, Markus Langer, Veronika Lazar, Astrid Schomäcker, Timo Speith, and Sarah Sterz.
\newblock On the relation of trust and explainability: Why to engineer for trustworthiness.
\newblock In {\em 2021 IEEE 29th International Requirements Engineering Conference Workshops (REW)}, pages 169--175, 2021.

\bibitem{trustworthy-ai-from-principles-to-practices}
Bo~Li, Peng Qi, Bo~Liu, Shuai Di, Jingen Liu, Jiquan Pei, Jinfeng Yi, and Bowen Zhou.
\newblock Trustworthy ai: From principles to practices.
\newblock {\em ACM Comput. Surv.}, 55(9), jan 2023.

\bibitem{mythos-of-model-interpretability}
Zachary~C. Lipton.
\newblock The mythos of model interpretability: In machine learning, the concept of interpretability is both important and slippery.
\newblock {\em Queue}, 16(3):31–57, jun 2018.

\bibitem{wordnet}
George~A. Miller.
\newblock Wordnet: A lexical database for english.
\newblock {\em Commun. ACM}, 38:39--41, 1995.

\bibitem{explanation-insights-from-the-social-sciences}
Tim Miller.
\newblock Explanation in artificial intelligence: Insights from the social sciences.
\newblock {\em Artificial Intelligence}, 267:1--38, 2019.

\bibitem{quantitative-evaluation-of-machine-learning-explanations}
Sina Mohseni, Jeremy~E Block, and Eric Ragan.
\newblock Quantitative evaluation of machine learning explanations: A human-grounded benchmark.
\newblock In {\em Proceedings of the 26th International Conference on Intelligent User Interfaces}, IUI '21, page 22–31, New York, NY, USA, 2021. Association for Computing Machinery.

\bibitem{comparing-automatic-and-human-evaluation-of-local-explanations}
Dong Nguyen.
\newblock Comparing automatic and human evaluation of local explanations for text classification.
\newblock In Marilyn Walker, Heng Ji, and Amanda Stent, editors, {\em Proceedings of the 2018 Conference of the North {A}merican Chapter of the Association for Computational Linguistics: Human Language Technologies, Volume 1 (Long Papers)}, pages 1069--1078, New Orleans, Louisiana, June 2018. Association for Computational Linguistics.

\bibitem{lime}
Marco~T{\'{u}}lio Ribeiro, Sameer Singh, and Carlos Guestrin.
\newblock "{W}hy should {I} trust you?": Explaining the predictions of any classifier.
\newblock {\em CoRR}, abs/1602.04938, 2016.

\bibitem{explainable-ai-meta-survey}
Waddah Saeed and Christian Omlin.
\newblock Explainable ai (xai): A systematic meta-survey of current challenges and future opportunities.
\newblock {\em Knowledge-Based Systems}, 263:110273, 2023.

\bibitem{quantifying-interpretability-and-trust}
Philipp Schmidt and Felix Biessmann.
\newblock Quantifying interpretability and trust in machine learning systems.
\newblock {\em ArXiv}, abs/1901.08558, 2019.

\bibitem{whose-right-reasons}
Terne~Sasha Thorn~Jakobsen, Laura Cabello, and Anders S{\o}gaard.
\newblock Being right for whose right reasons?
\newblock In Anna Rogers, Jordan Boyd-Graber, and Naoaki Okazaki, editors, {\em Proceedings of the 61st Annual Meeting of the Association for Computational Linguistics (Volume 1: Long Papers)}, pages 1033--1054, Toronto, Canada, July 2023. Association for Computational Linguistics.

\bibitem{Trustworthy-Explainability-Acceptance-Food}
Suleyman Uslu, Davinder Kaur, Samuel~J. Rivera, Arjan Durresi, Mimoza Durresi, and Meghna Babbar‐Sebens.
\newblock Trustworthy acceptance: A new metric for trustworthy artificial intelligence used in decision making in food-energy-water sectors.
\newblock In {\em International Conference on Advanced Information Networking and Applications}, 2021.

\bibitem{attention-is-not-not-explanation}
Sarah Wiegreffe and Yuval Pinter.
\newblock Attention is not not explanation.
\newblock In {\em Conference on Empirical Methods in Natural Language Processing}, 2019.

\bibitem{trustworthy-ai}
Jeannette~M. Wing.
\newblock Trustworthy ai.
\newblock {\em Commun. ACM}, 64(10):64–71, sep 2021.

\bibitem{evaluating-the-quality-of-machine-learning-explanations}
Jianlong Zhou, Amir~Hossein Gandomi, Fang Chen, and Andreas Holzinger.
\newblock Evaluating the quality of machine learning explanations: A survey on methods and metrics.
\newblock {\em Electronics}, 2021.

\bibitem{class-noise-vs-attribute-noise}
Xingquan Zhu and Xindong Wu.
\newblock Class noise vs. attribute noise: a quantitative study of their impacts.
\newblock {\em Artif. Intell. Rev.}, 22(3):177–210, nov 2004.

\end{thebibliography}
\bibliographystyle{plain}

\vspace{12pt}

\end{document}